\documentclass[letterpaper,8pt]{extarticle}  % "extarticle" gives more global font size options. use 8pt with Montserrat and 10pt with EBGaramond
\usepackage{import}
\usepackage{local}
\usepackage{tipa}
\usepackage{soul}
\usepackage{color}
\usepackage{placeins}

\usepackage[left=.65in,top=.65in,bottom=.65in,right=.65in]{geometry}
\title{Covertly improving intelligibility with data-driven adaptations of speech timing}

\author{%
\textbf{Paige Tuttösí\textcolor{Accent}{\textsuperscript{1,2,3,}}, %
Angelica Lim\textcolor{Accent}{\textsuperscript{1}}, %
H. Henny Yeung\textcolor{Accent}{\textsuperscript{3}}, %
Yue Wang\textcolor{Accent}{\textsuperscript{3}}, %
Jean-Julien Aucouturier\textcolor{Accent}{\textsuperscript{2}}%
}\\
\begin{small}\textcolor{Accent}{\textsuperscript{1}}School of Computing Science, Simon Fraser University, Canada \\
\textcolor{Accent}{\textsuperscript{2}}Université Marie et Louis Pasteur, SUPMICROTECH, CNRS, institut FEMTO-ST, France \\
\textcolor{Accent}{\textsuperscript{3}}Department of  Linguistics, Simon Fraser University, Canada \\
\end{small}
}

\date{}
\begin{document}
\maketitle
\thispagestyle{empty}

%%%%%%%%%%%%%%%%%%%%%%%%%%%%%%%%%%%%%%%%%%%%%%%%%%%%%%%%%%%%%%%%%%%%%
% Abstract
%%%%%%%%%%%%%%%%%%%%%%%%%%%%%%%%%%%%%%%%%%%%%%%%%%%%%%%%%%%%%%%%%%%%%
\section{Abstract}

\begin{doublespacing}
%\begin{linenumbers}

\noindent
\textbf{\textcolor{Accent}{Human talkers often address listeners with language-comprehension challenges, such as hard-of-hearing or non-native adults, by globally slowing down their speech. However, it remains unclear whether this strategy actually makes speech more intelligible. Here, we take advantage of recent advancements in machine-generated speech allowing more precise control of speech rate in order to systematically examine how targeted speech-rate adjustments may improve comprehension. We first use reverse-correlation experiments to show that the temporal influence of speech rate prior to a target vowel contrast (ex. in English, the tense–lax distinction; \textit{peel}–\textit{pill}) in fact manifests in a scissor-like pattern, with opposite effects in early versus late context windows; this pattern is remarkably stable both within individuals and across native L1-English listeners and L2-English listeners with French, Mandarin, and Japanese L1s. Second, we show that this speech rate structure not only facilitates L2 listeners’ comprehension of the target vowel contrast, but that native listeners also rely on this pattern in challenging acoustic conditions. Finally, we build a data-driven text-to-speech (TTS) algorithm that replicates this temporal structure on novel speech sequences. Across a variety of sentences and vowel contrasts, listeners remained unaware that such targeted slowing improved word comprehension. Strikingly, participants instead judged the common strategy of global slowing as clearer, even though it actually increased comprehension errors. Together, these results show that targeted adjustments to speech rate significantly aid intelligibility under challenging conditions, while often going unnoticed. More generally, this paper provides a data-driven methodology to improve the accessibility of machine-generated speech which can be extended to other aspects of speech comprehension and a wide variety of listeners and environments.}}

\section{Introduction}

A classical view of speech perception is that it is a local process in which sounds are compared to stored auditory representations \citep{STRAN89}. In vowel perception, these representations are often spectral, such as templates in formant space \citep{PET52, SYR86}. Speech sounds, however, are highly variable not only across speakers but also within connected speech, and it is now widely documented that their perception necessarily operates relative to their acoustic context \citep{STIL20}. For instance, following a phrase where the first formant ($F_1$) is shifted down toward lower frequencies, a target vowel /\textipa{I}/ may be perceived as the higher-$F_1$ /\textipa{E}/, even when its local acoustical content does not change \cite{LADE57,STIL18}. 

Among several contextual influences, \emph{speech rate effects}, where listeners use speech rate information in a preceding sentence to discriminate subsequent phonemes and words, have become a hallmark of speech perception research \cite{MillerLiberman1979, KIDD89,SommersNygaardPisoni1994, NEWM96,STIL20}. In laboratory-based studies, speech rate effects (also called ``rate-dependent speech perception'' \cite{Miller1987, REIN22}) have been documented in several languages and listening conditions, and are thought to be a core component of how one processes the temporal contrasts that are essential to speech comprehension. These contrasts include not only vowel length distinctions in languages that use duration as a primary cue (e.g. word pairs like /\textipa{k\'ado}/, \emph{corner} vs /\textipa{k\'a:do}/, \emph{card} in Japanese; \cite{Hirata2004}) or secondary cue (e.g., tense vs lax vowels in English such as \emph{peel}: /\textipa{i}/ vs \emph{pill}: /\textipa{I}/), but also the voice onset time of consonants \cite{KIDD89} and contrasts in formant duration such as \textipa{[ba]} vs \textipa{[wa]} \cite{WADE05}. The cognitive characteristics of such effects are also well-studied, both in terms of chronometry \cite{MASLO20}, domain-specificity \cite{PITT16} and automaticity \cite{BOSK20}. 

Yet, despite a wealth of laboratory studies, the causal influence of speech rate effects in the comprehension of ecological speech (i.e., speech with diverse lexical content and in varied listening conditions) has been difficult to establish, and may be less significant than estimated in the lab \cite{BOSK20}. First, most studies attempting to reproduce context effects in more ecological settings, such as in background noise with reverberant acoustics \cite{REIN22} or with additional speaker variability \cite{BOSK20}, have generally found smaller effects than estimated in clearer conditions, raising questions about whether strategically manipulating speech rate in everyday speech can actually facilitate or impede its clarity. Second, speech-rate effects observed in the laboratory strikingly contradict the pragmatics of how fluent speakers manipulate speed to optimize speech comprehension. For instance, when speakers attempt to direct `clear speech' to hard-of-hearing or non-native listeners, they overwhelmingly do this by slowing down their production regardless of phonemic content \cite{SMIL09,AOKI24} -- a strategy that context research predicts to be maladaptive, as it would systematically bias perception towards shorter over longer sounds. Third, the exact temporal mechanisms by which the speech perception system may take in rate information remains mostly unknown when extended to sentences of arbitrary length. Depending on the type of laboratory task and/or stimuli, the observed influence of speech rate on subsequent phoneme distinction has either been shown to accumulate over several sentences \cite{JOHN99}, or to be limited to a temporal window of one or two adjacent phonemes \cite{Reinisch13}. Indeed, there is likely a hierarchy of `distal' (i.e. earlier, longer-term) and `proximal' (i.e. immediately preceding) effects (possibly competing) \cite{STIL20} but, while these effects may be easy to induce in short controlled sentences, it is unclear how they can be generalized to arbitrary sentences with non-stereotyped syntax and multiple loci of interest \cite{KRAUS02}. In short, we are lacking an 'algorithmic theory' of how rate information is used to make speech clearer. 

\begin{figure*}[t]
  \centering
  \includegraphics[width=\linewidth]{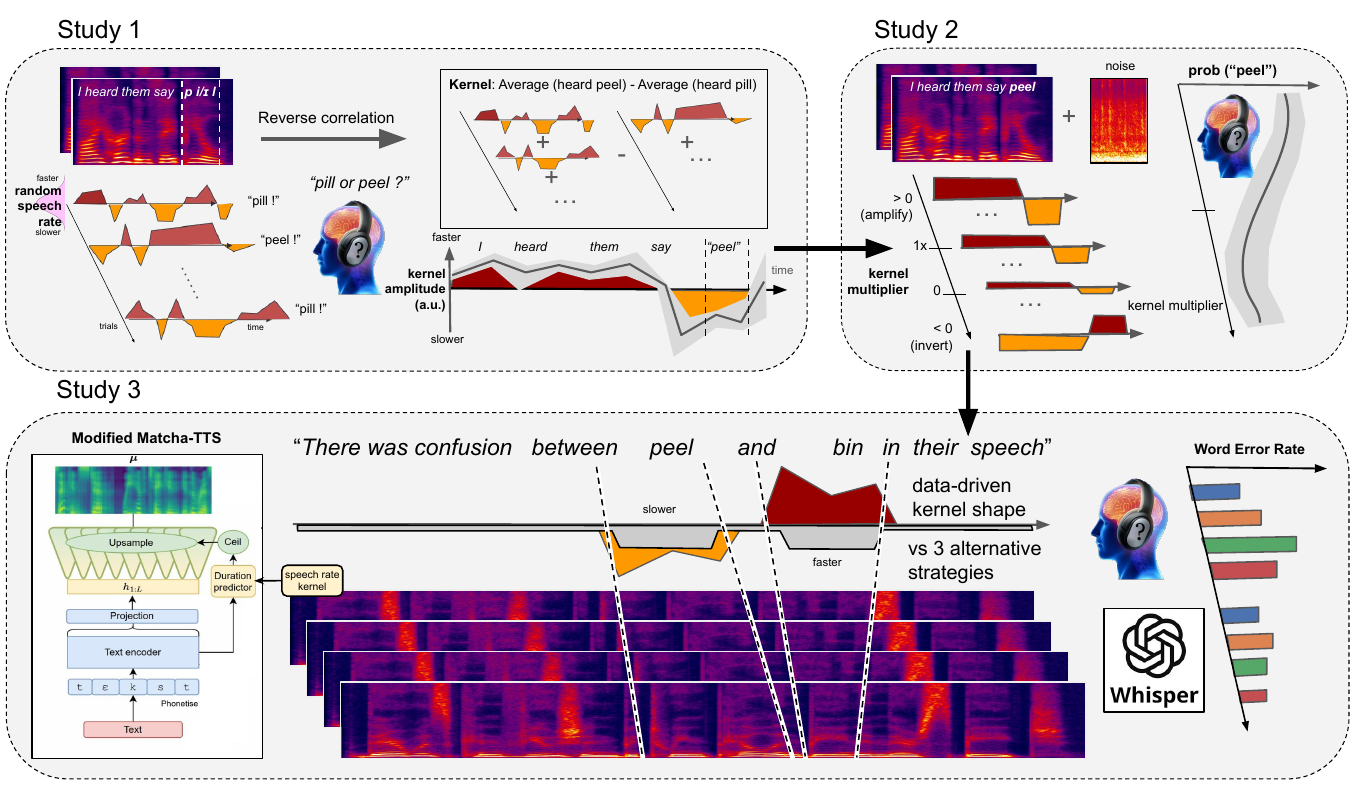}
  \caption{{\bf A data-driven algorithm to manipulate speech rate in clear speech:} In a series of three studies, we uncover what parametric temporal contours of speech rate is causal for improving word recognition, in a way that can then be ‘synthesized’ to generate novel, machine-generated speech. {\bf (Top-left) } Using a speech signal-processing technique, we systematically varied speech rate in phrases surrounding difficult word contrasts (e.g. pill vs peel), and used reverse correlation to extract the temporal contour of rate (or `kernel') that biases word recognition in one or the other direction (Study 1). {\bf (Top-right) } We then resynthesized short phrases with graded intensity of speech rate manipulation (`kernel multiplier'), to quantify the kernel effect on word comprehension and investigate the effect of background noise (Study 2). {\bf (Bottom) } Finally, we combined the data-driven kernel shape and multiplier with state-of-the-art text-to-speech (TTS) synthesis to provide an algorithmic way to make synthesized speech clearer, and tested its effect on both human and machine listeners (Study 3).}
  \label{fig:procedure}
  %\vspace*{-5mm}
\end{figure*}

%\FloatBarrier

Determining the temporal characteristics of information intake during the perception of arbitrary connected speech is methodologically challenging. While hypothesis-driven experiments can establish the causal influence of specific cues on word contrasts by systematically varying their magnitude (e.g. 13 distal speech rates on the perception of a target word in \cite{HEFF13}), simultaneously assessing the relative weight of several temporal regions of interest becomes quickly impractical \cite{NEWM96}. To document such temporal dynamics, several studies have relied on eye-tracking in visual search tasks (e.g. printed words), and compared the time course of eye fixations to the occurrence of contextual cues \cite{Reinisch13, SHATZ06}. However, this type of information does not directly translate to a causal mechanism that can be implemented and tested on listeners, especially across multiple languages or language proficiencies. A key breakthrough would be to to determine what parametric temporal contours of rate information may be causal for speech perception, in a way that can then be algorithmically `synthesized' to generate novel speech.

Here, we propose a data-driven route towards building an algorithm to manipulate the temporal structure of rate in arbitrary connected speech in a way that maximizes its clarity for listeners. Using a series of reverse-correlation experiments, we first extract the exact temporal weighting of speech rate information in short phrases and test its preservation across native and non-native listeners of four different languages (Study 1). We then show that this speech-rate structure not only facilitates L2 listeners’ comprehension of vowel contrast, but that native listeners also rely on this pattern in challenging acoustic conditions (Study 2). Finally, using state-of-the-art speech synthesis, we provide an algorithmic way to enforce that temporal structure on arbitrary speech (Study 3) and show that, strikingly, listeners remain unaware of the facilitating effect of targeted speech
rate on word comprehension. Taken together, these results provide a major step towards ecological speech synthesis that is adaptive to listener differences. 

\section*{Results}

\subsection*{Study 1: The temporal contour of rate information intake is remarkably stable across individuals and languages}

Using a series of reverse-correlation experiments, we first extracted the temporal contour of rate information intake within isolated words (Study 1a) and in a phrase leading to a target word (Study 1b\&c). To do so, we used a speech signal-processing technique (\cite{BURR19}, see \emph{Methods}) to systematically and concurrently vary pitch (F0) and speech rate in English and French phrases surrounding ambiguous pairs of vowels that are difficult for L2 speakers: English \emph{pill}: /\textipa{I}/ vs \emph{peel}: /\textipa{i}/, and French \emph{pull} (sweater): /\textipa{y}/ vs \emph{poule} (chicken): /\textipa{u}/. We then asked several groups of L1 and L2 speakers (French-L1, English-L2: \textit{N} = 54; English-L1, French-L2: \textit{N} = 60; Mandarin-L1, English-L2: \textit{N} = 30; Japanese-L1, English-L2: \textit{N} = 16) to categorize these manipulated sentences as containing one or the other vowel. From their decisions, we then used reverse correlation to reconstruct, in a purely data-driven manner, the prosodic profiles that bias the perception of these word pairs (Fig. \ref{fig:procedure}, top left). 

\subsubsection*{Study 1a -- Isolated French and English words, French/English L1 and L2 listeners}
To validate the procedure, we first asked whether a reverse-correlation procedure can recover informative patterns of speech rate in isolated words, presented without a sentence context. 

In French, we selected the word pair \emph{pull} (sweater) and \emph{poule} (chicken), which features a high-front /\textipa{y}/ vs high-back /\textipa{u}/ vowel contrast often reported to be difficult for English-L1 speakers \cite{LEVY10}. We presented \textit{N} = 25 French-L1 speakers and \textit{N} = 25 English-L1, French-L2 speakers with 250 trials consisting of an hybrid 250-ms word \emph{pull/poule}, manipulated with formant morphing to be perceptually intermediate between /\textipa{y}/ and /\textipa{u}/, and in which we randomly varied the pitch and speech rate on the four successive 100-ms windows. For each trial, participants were asked which of two target words they recognized. We then used the `classification-image' method \citep{Mur11} to compute first-order \emph{kernels} (see \emph{Methods}) representing what rate transformation should be applied to the base word in order to increase the probability of recognizing one target word or another. Because vowel duration isn't considered a primary cue for the French /\textipa{y}-\textipa{u}/ contrast \cite{OSHAU81}, there was no clear phonological prediction for how speech rate would influence its perception. Negative values for the \emph{pull} kernel (Supplementary Fig. 1-left) revealed that its perception was facilitated by a shortening of the sound, and conversely positive values in the \emph{poule} kernel corresponded to a lengthening of the sound (0.1s: $t(25)=2.80$,$p=.010$; 0.2s: $t(25)=3.52$, $p=.002$). The same pattern was found for French-L2 speakers  (0.1s: $t(25)=4.57$, $p<.001$; 0.2s: $t(25)=2.73$, $p=.012$). 

In English, we selected the word pair \emph{pill} and \emph{peel}, which features the tense-lax vowel contrast /\textipa{I}-\textipa{i}/, the former being articulated with greater muscular effort, slightly higher tongue position, and longer durations. As above, this contrast is known to be difficult for French-L1 speakers \cite{IVE07}. We presented \textit{N} = 25 English-L1 speakers and \textit{N} = 25 French-L1, English-L2 speakers with manipulated hybrids of \emph{pill/peel}. Because duration is a secondary cue in English tense/lax contrasts \cite{SOLE10}, the phonological prediction was that the perception of \emph{peel} would be driven by slower speech rate than \emph{pill}. While no such difference was found for English-L1 speakers, French-L1 speakers did display the expected faster rate for /\textipa{I}/ (Supplementary Fig. 1-right; 0.2s: $t(25)=3.13$, $p=.005$). 

In short, despite some variations linked to language proficiency, we validated that our reverse-correlation procedure was able to uncover speech rate characteristics of vowels both in phonetically predictable and in undocumented cases, when presented in isolated words in French and English.

\begin{figure*}[t]
  \centering
  \includegraphics[width=\linewidth]{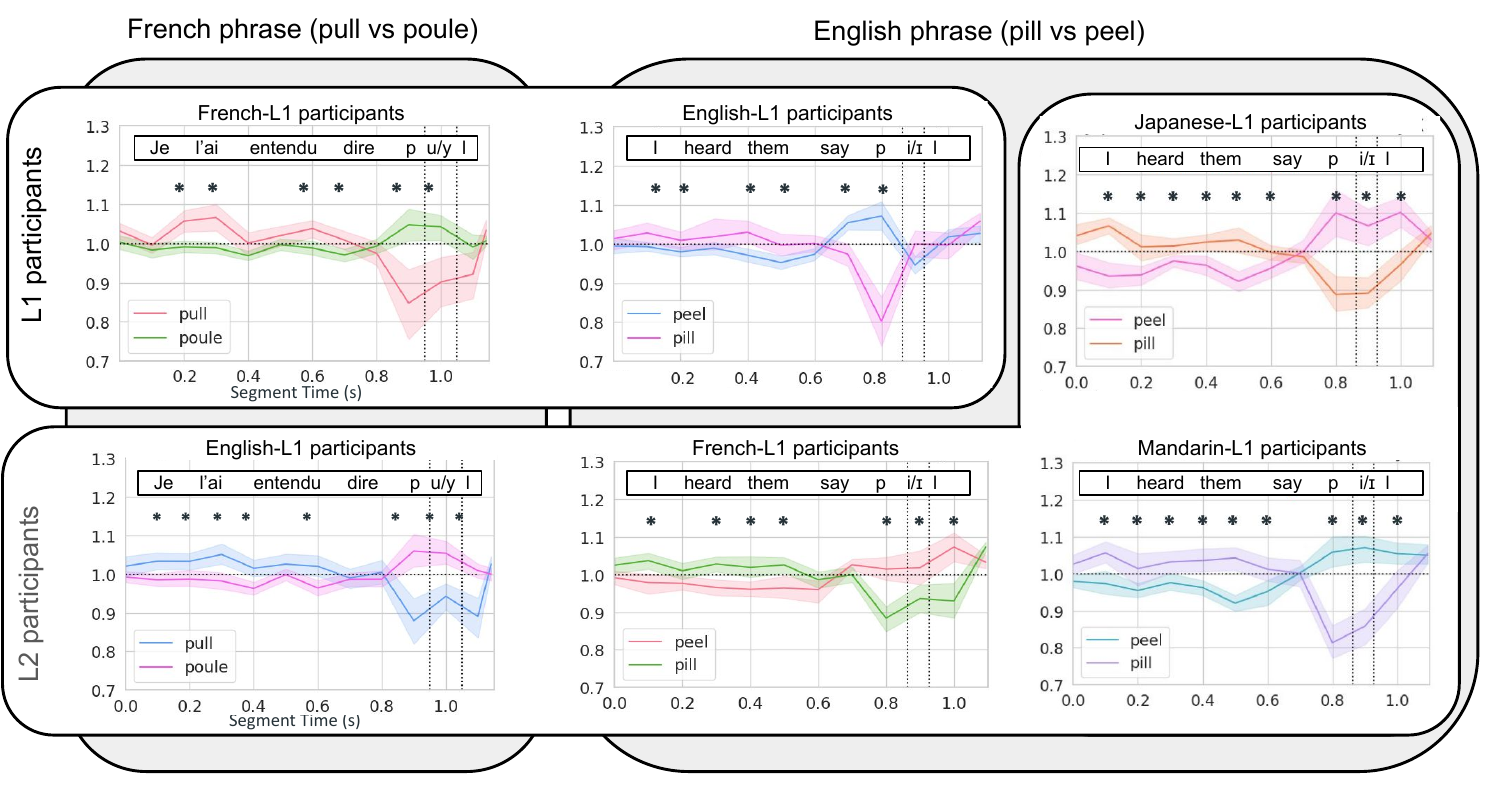}
  \caption{ {\bf The temporal contour of rate information intake in leading phrasal context is remarkably stable across individuals and languages (study 1b). } We systematically varied speech rate in sentences preceding words distinguished by pairs of vowels that are difficult for L2 speakers: French \emph{pull} (sweater: /\textipa{y}/) vs \emph{poule} (chicken: /\textipa{u}/); English pill (/\textipa{I}/) vs peel (/\textipa{i}/). Reverse correlation kernels showed a clear `` scissor-shape'' pattern, in which slower speech 800-300ms pre-target biased the perception of subsequent words in the direction of the phonetically \emph{faster} option (French: \emph{pull}, left; English: \emph{pill}, middle \& right), and slower speech starting 100-200ms pre-target biased the immediately following sound in the direction of the phonetically \emph{slower} option (French: \emph{poule}; English: \emph{peel}). This pattern was conserved almost identically across four languages in both L1 (Top-Left: French-L1 on French; Top-Middle: English L1 on English) and L2 participants (Bottom-Left: English-L1 on French; Bottom-Middle: French L1 on English; Right: Japanese- and Mandarin Chinese-L1 on English). }
  \label{fig:revcor}
  %\vspace*{-5mm}
\end{figure*}

\subsubsection*{Study 1b -- French and English sentences, French/ English L1 and L2 listeners}

We then used the same reverse-correlation procedure to extract the temporal contour of rate information intake over a phrasal context preceding a target word. We embedded \emph{pill/peel} and \emph{pull/poule} hybrids in the final position of a \textit{M} = 1.1s non-informative sentence (English words: \emph{``I heard them say XXX''}; French words: \emph{``Je l'ai entendu dire XXX''}), and randomly manipulated the pitch and speech rate of the resulting stimulus in 13 successive 100-ms time windows. 

As before, we then asked \textit{N} = 25 English-L1, French-L2 and \textit{N} = 25 French-L1, English-L2 speakers to listen and categorize 250 random trials as one or the other word alternative, and computed reverse-correlation kernels that described what speech rate transformation should be applied (and in what time window) to the leading phrasal context in order to bias the perception of the final target word. In addition to reproducing previous results within the target word (faster s/\textipa{y}/ in French-L1 speakers, 1.0s: $t(25)=2.99$, $p=.006$ and English-L1 speakers, 1.0s: $t(25)=4.04$, $p<.001$, 1.1s: $t(25)=3.71, p=.001$; shorter /I/ in French-L1 speakers, 1.0s: $t(25)=3.61$, $p=.001$), the kernels of both English and French-L1 participants showed a clear ``scissor-shape'' pattern (i.e. opposing effects crossing over before the target word, see Figure \ref{fig:revcor}, top left and middle). This effect, which spanned all the leading phrasal context before the target word, consisted of two components: First, a ``distal'' contrastive effect starting as early as 800ms pre-target and extending up to 300ms pre-target, in which slower speech biased the perception of the subsequent word in the direction of the vowel that is often found, phonetically, to be 'faster' (French \emph{pull}: 0.2s, $t(25)=3.72$, $p=.001$, 0.3s: $t(25)=3.86$, $p=.001$, 0.6s: $t(25)=2.89$, $p=.008$, 0.7s: $t(25)=2.24$, $p=.01$; English \emph{pill}:  0.1s: $t(25)=3.35$, $p=.003$, 0.2s: $t(25)=2.53$, $p=.018$, 0.3s: $t(25)=4.12$, $p<.001$, 0.4s: $t(25)=2.99$, $p=.006$, 0.6s: $t(25)=3.46$, $p=.002$); and second, a ``proximal'' congruent effect starting 100-200ms pre-target, in which slower speech biased the immediately following sound in the direction of the vowel that is often found, phonetically, to be 'slower' (French \emph{poule}: 0.9s: $t(25)=-3.25$, $p=.003$; English \emph{peel}: 0.9s: $t(25)=-3.76$, $p=.001$). 

Remarkably, that scissor-shape pattern of contextual rate influence was conserved almost identically in all groups of L2 participants. Tested on English sentences, French-L1 participants showed the same kernel as English-L1 participants (Figure \ref{fig:revcor}, bottom middle), with distal contrastive effects (0.1s: $t(25)=-3.02$, $p=.006$,
0.3s: $t(25)=-3.53$, $p=.002$, 0.4s: $t(25)=-3.07$, $p=.005$, 0.5s: $t(25)=-2.75$, $p=.011$), and a strong proximal congruent effect (0.8s: $t(25)=4.25$, $p<.001$). Conversely, tested on French sentences, English-L1 participants also showed the same distal (0.1s: $t(25)=-
3.35$, $p=.003$, 0.2s: $t(25)=-2.53$, $p=.018$, 0.3s: $t(25)=-4.12$, $p<.001$, 0.4s: $t(25)=-2.99$, $p=.006$, 0.6s: $t(25)=-3.46$, $p=.002$) and proximal effects (0.9s: $t(25)=3.76$, $p=.001$) as French-L1 participants (Figure \ref{fig:revcor}, bottom left).

\subsubsection*{Study 1c - English sentences, English-L2, Mandarin and Japanese-L1 listeners}

To test whether this pattern of results extended to a larger group of non-native L1s, we tested two additional groups of English-L2 speakers (Mandarin-L1: N=25; Japanese-L1: N=19) on the same task. These two languages differ in how speech rate is involved in L1 experience with vowel properties: rate is a primary phonemic cue for vowel contrast in Japanese \citep{Hirata2004}, whereas pitch is the primary cue in Mandarin (see {\bf Discussion}). Tested in English, both groups again showed strikingly similar scissor patterns as English-L1 and French-L1 speakers (Mandarin-L1: 0.1s: $t(25)=-3.39$, $p=.002$, 0.2s: $t(25)=- 2.11$, $p=.044$, 0.3s: $t(25)=-3.01$, $p=.006$, 0.4s: $t(25)=-3.65$, $p=.001$, 0.5s: $t(25)=-5.66$, $p<.001$, 0.6s: $t(25)=-2.28$, $p=.031$, 0.8s: $t(25)=6.36$, $p<.001$, 0.9s: $t(25)=5.43$, $p<.001$, 1.0s: $t(25)=2.44$, $p=.022$ -- Figure \ref{fig:revcor}, top right; Japanese-L1: 0.1s: $t(13)=-5.25$, $p<.001$, 0.2s: $t(13)=-2.82$, $p=.014$, 0.3s: $t(13)=-2.63$, $p=.021$, 0.4s: $t(13)=-3.02$, $p=.010$, 0.5s: $t(13)=-4.51$, $p=.001$, 0.6s: $t(13)=-2.14$, $p=.052$, 0.8s: $t(13)=4.01$, $p=.001$, 0.9s: $t(13)=4.08$, $p=.001$, 1.0s: $t(13)=3.39$, $p=.005$ -- Figure \ref{fig:revcor}, bottom right).

%PUT IN SUPPLEMENTARY Finally, because we simultaneously manipulated both pitch and speech-rate in the above reverse-correlation experiments (see \emph{Methods}), we re-analysed participant responses to compute pitch kernels describing whether, independently of speech rate,  higher or lower pitch in context sentences biased the perception of the target words. Contrary to speech-rate, the temporal contour of pitch information in-take in sentences preceding the target words appeared both less robust and largely language-dependent (see Supplementary material). In French stimuli, contextual pitch was not associated with any contrastive effect, but rather a small proximal congruent effect 200-
%300ms pre-target (French-L1, 0.7s: t(25)=-2.18, p=.040; English-L1, 0.8s: t(25)=-4.05, p<.001), i.e. an increase of pitch immediately before the target word biased the response towards the higher-pitch alternative /\textipa{y}/ (Figure \ref{fig:S2}-where). In English stimuli, pitch showed contrastive effects within the phrase, both distally (0.1s: t(25)=3.23, p=.003, 0.3s: t(25)=3.68, p=.001) and in the immediate proximity of the target word (0.8s: t(25)=2.91, p=.007) (Figure \ref{fig:S2}-where), but these effects were largely absent in the three groups of English-L2 speakers. In sum, while speech-rate effects were remarkably conserved across listeners and languages, pitch does not seem to have a consistent contextual effect on the two vowel contrasts tested here. 

\subsection*{Study 2: How speech-rate scissor structures affect word error rate in native and non-native listeners}

To examine the practical significance of the reverse-correlation kernels found in Study 1, as well as test their generalization to other English vowel pairs where duration is expected to be a (secondary) cue (i.e., tense/lax vowels), we applied the scissor-shape kernel to a new set of English sentences (\emph{I heard them say [peel, pill,fool,full]}) in which, in contrast to Study 1, the target words were not manipulated to have ambiguous formants. For each sentence, we applied the scissor-shape rate manipulation at 11 levels of positive and negative (i.e., reversed) intensity, corresponding to context speed multipliers ranging from 0.67x to 1.5x by increments of 0.1 (see \emph{Methods}), as well as a non-manipulated version at 1x speed. We then asked four groups of participants to categorize the target word in its two alternatives (\emph{pill/peel} and \emph{full/fool}), and analyzed recognition accuracy as a function of manipulation intensity. 

\subsubsection*{Study 2a - Speech-rate scissor structures affect L2 English vowel comprehension across three L1s}

We first validated whether non-manipulated stimuli were correctly recognized by all participant groups. English-L1 participants (\textit{N} = 25) had almost perfect accuracy for non-manipulated sentences, at 96\% for \emph{pill}, and 94.4\% for \emph{peel} and \emph{full}. However, for \emph{fool} specifically, our procedure for stimulus generation (see \emph{Methods}) failed to synthesize clear formants, and native speakers struggled to identify the correct word even when non-manipulated (accuracy 32\%). All English-L2 participant groups had relatively high recognition accuracy for non-manipulated sentences, except for the same \emph{fool} (\textit{N} = 25 French-L1: 80.8\% accuracy for \emph{pill}, 76.8\% for \emph{peel}, 83.2\% for \emph{full}, 32\% for \emph{fool}; \textit{N} = 10 Mandarin-L1: 88.0\% accuracy for \emph{pill}, 82.5\% for \emph{peel}, 78.0\% for \emph{full}, 17.0\% for \emph{fool}; \textit{N} = 10 Japanese-L1: 98.0\% accuracy for \emph{pill}, 88.0\% for \emph{peel}, 98.0\% for \emph{full}, 28.0\% for \emph{fool}). 

\begin{figure*}[t]
  \centering
  \includegraphics[width=\linewidth]{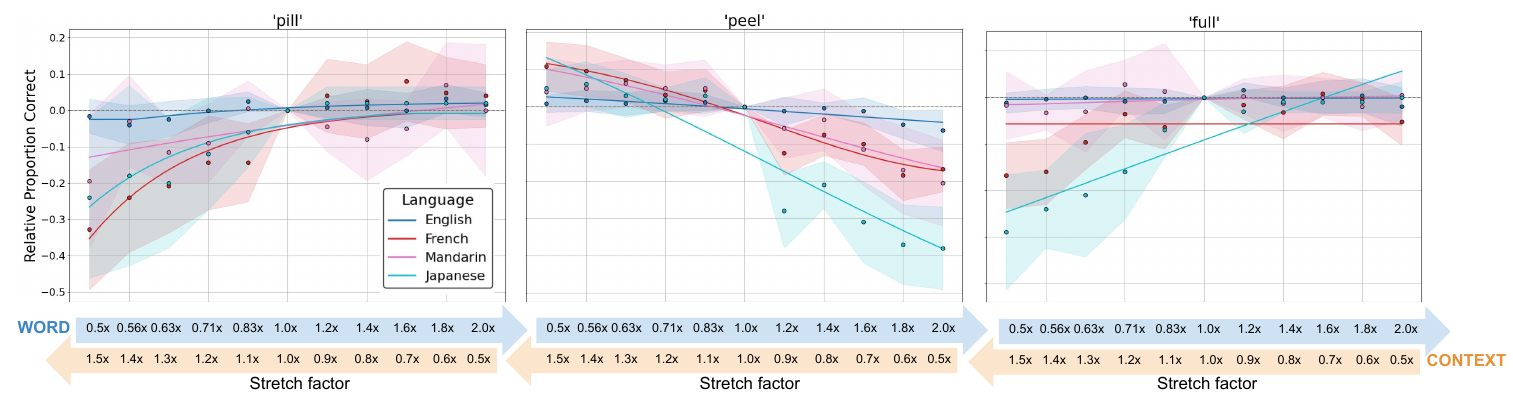}
  \caption{ {\bf Speech-rate scissor structures affect L2 English vowel comprehension across three L1s (Study 2a)} Recognition accuracy for English sentences (\emph{I heard them say pill/peel/full}) where the scissor-shape rate manipulation was applied at 11 levels of positive and negative (i.e., reversed) intensity, corresponding to distal `context' speed multipliers ranging left to right from 1.5x (faster) to 0.67x (slower) and, simultaneously, proximal `word' multipliers from 0.5x (slower) to 2.0x (faster). In all three English-L2 groups (Red: French; Pink: Mandarin; Light Blue: Japanese), accuracy was significantly swayed by applying the speech-rate kernel in one or the other direction. In contrast, English-L1 participants (blue) remained insensitive to the kernel manipulations for all three words. Accuracy displayed normalized with respect to 1x speed. Solid lines: fitted logistic regression. Shaded areas: 95\% confidence interval on average accuracy. }
  \label{fig:validation_1}
  %\vspace*{-5mm}
\end{figure*}

We then examined the effect of applying the scissor-shape kernel of speech rate at different levels of magnitude. In all three English-L2 groups, percentage accuracy was significantly swayed by applying the speech-rate kernel in one or the other direction, resulting in characteristic sigmoid shapes fitted by generalized logistic models (Figure \ref{fig:validation_1}). For instance, in French-L1 speakers (\textit{N} = 25), increasing speech rate at 0.67x duration distally while decreasing speech rate at 2.0x duration proximally increases the proportion correct for \emph{peel} by more than (absolute) 20\% (baseline: 76.8\%, manipulated: 98.4\%, $t(24)=3.36$, $p=.02$), while the opposite transformation decreases accuracy by about 30\% (manipulated: 43.2\%, $t(24)=-5.74$, $p<.001$; Figure \ref{fig:validation_1}-colour-where). Both words of the same vowel type had similar directional effects (tense \emph{pill}/\emph{full}: upwards right; lax \emph{peel}/\emph{fool}: upwards left). Similar trends were observed in Mandarin-L1 (Figure \ref{fig:validation_1}-colour) and Japanese-L1 effects (Figure \ref{fig:validation_1}-colour). 

In short, manipulations of speech rate in distal and proximal context had a potentially large facilitating or disrupting effect on word comprehension for English-L2 speakers, to the point of reaching perfect accuracy when rate is manipulated in the appropriate direction (or chance level, when manipulated in the opposite direction). 

\subsubsection*{Study 2b: Native listeners do not primarily rely on speech rate, but use it as a fall-back strategy in challenging acoustic conditions}

English-L1 results were in stark contrast with English-L2, because these participants remained remarkably insensitive to the kernel manipulations for all three words where performance was at ceiling (\emph{peel}, \emph{pill} and \emph{full}) (Figure \ref{fig:validation_1}-colour). However, strikingly, when tested on \emph{fool}, a sound which was generated unintentionally ambiguously, English-L1 speakers showed L2-like behaviour, with recognition accuracy increasing as distal speech rate increased and proximal rate decreased (Figure \ref{fig:validation_1}-colour-where): (baseline: 32\%; distal 0.67x duration, proximal 2.0x duration: 66.4\%, $t(24)=5.58$, $p=.001$). This suggested that, while L1-speakers do not typically rely on speech-rate to facilitate word perception, they may start using duration cues as a fall-back strategy in situations where timbral information is ambiguous or masked. 

\begin{figure*}[t]
  \centering
  \includegraphics[width=\linewidth]{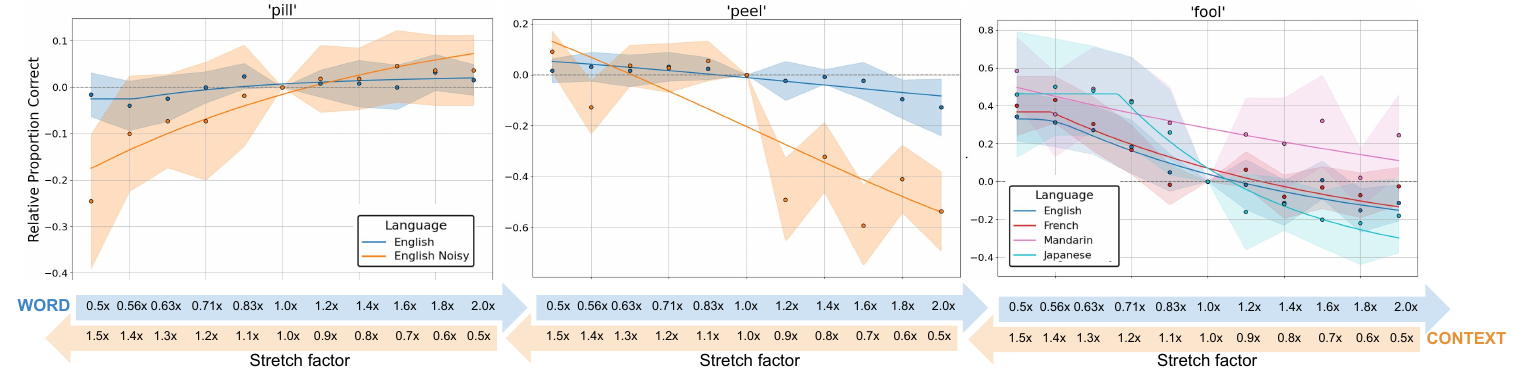}
  \caption{ {\bf Native listeners use speech rate as a fall-back strategy in challenging acoustic conditions (study 2b) } English-L1 recognition accuracy for manipulated English sentences (\emph{I heard them say pill/peel/fool}). English-L1 participants showed L2-like behaviour only in the presence of background noise (orange. Left: \emph{pill}; Middle: \emph{peel}; for both: blue curve same as Fig. \ref{fig:validation_1}) or when the sound was synthesized with ambiguous formants (Right: \emph{full}). Rate manipulation levels similar as Fig. \ref{fig:validation_1}. Accuracy displayed normalized with respect to 1x speed. Solid lines: fitted logistic regression. Shaded areas: 95\% confidence interval on average accuracy.}
  \label{fig:validation_noise}
  %\vspace*{-5mm}
\end{figure*}

To confirm this possibility, we tested another group of \textit{N} = 25 English-L1 speakers on stimuli manipulated with distortion, reverberation and background crowd noise (see \emph{Methods}). While baseline performance for these sounds remained high (84.5\% for \emph{pill}, 81.8\% for \emph{peel}), participants showed an effect of rate manipulation on word recognition, which mirrored the behaviour of English-L2 participants in non-noisy conditions (Figure \ref{fig:validation_noise}. This pattern of result is consistent with the idea that both types of speakers use duration as a cue, but with different weighting: native speakers who can easily use primary spectral cues to differentiate minimal pairs will use duration as a secondary cue and only use it in challenging listening conditions, while duration may be used as a primary cue by L2 speakers, who have more difficulty with a target spectral distinction. 

\subsection*{Study 3: A data-driven text-to-speech (TTS) algorithm that replicates this temporal structure on arbitrary speech sequences}

Taken together, results from Study 1-2 provide a promising strategy to improve the clarity of machine text-to-speech synthesis (TTS) for L2 speakers, by manipulating duration cues to enforce the correct perception of tense/lax alternatives. To provide a complete causal test of the effect of speech-rate changes on word error rate in fully connected speech, we built a custom text-to-speech synthesis model, based on the state-of-art MatchaTTS model \cite{mehta2024matcha}, which we extended to enforce the kernel's temporal structure on arbitrary speech. To do so, we created an algorithm which automatically parses input English text to identify words containing tense or lax vowels, and then applies appropriate duration changes to the context preceding these target sounds. 

To specify the contextual change, we used the kernels from Study 1-2 with the following adaptations: First, to avoid exaggerated boundary effects on preceding words, we restrict the algorithm to only use proximal context effects. For a tense vowel such as \emph{peel}, the algorithm synthesizes preceding speech with a gradually increasing stretch starting 6 phonemized items before target (approximately 200-300ms, per Study 1), peaking at 1.6x duration (per Study 2) on the entire word containing the vowel. Second, to avoid abrupt changes of rate immediately after the targeted vowel, we gradually decreased rate back to the base speech rate over the 6 phonemized items following the word. Finally, to avoid exaggerated contrasts of speed within pairs of sentences, we do not apply any stretch manipulation to lax vowels (e.g. \emph{pill}), based on the rationale that Study 2 has shown it easier to degrade the recognition of such sounds than to improve it (Figure \ref{fig:validation_1}-where). To validate that this strategy (proximal slowing-down for tense vowels, no change for lax vowels) does not overly impair the facilitation effect for non-native speakers, we replicated the same procedure as Study 2, this time only manipulating the rate transformation intensity of the proximal neighborhood of the word: tested on N=25 French-L1 speakers, the tense effect remained strong and significant, at about +20\% relative accuracy for \emph{peel} and +40\% accuracy for \emph{fool}, while the same effect on lax vowels was confirmed unnecessary as it did not significantly improve performance over baseline (Supplementary Fig. 2).

\begin{figure*}[t]
  \centering
  \includegraphics[width=\linewidth]{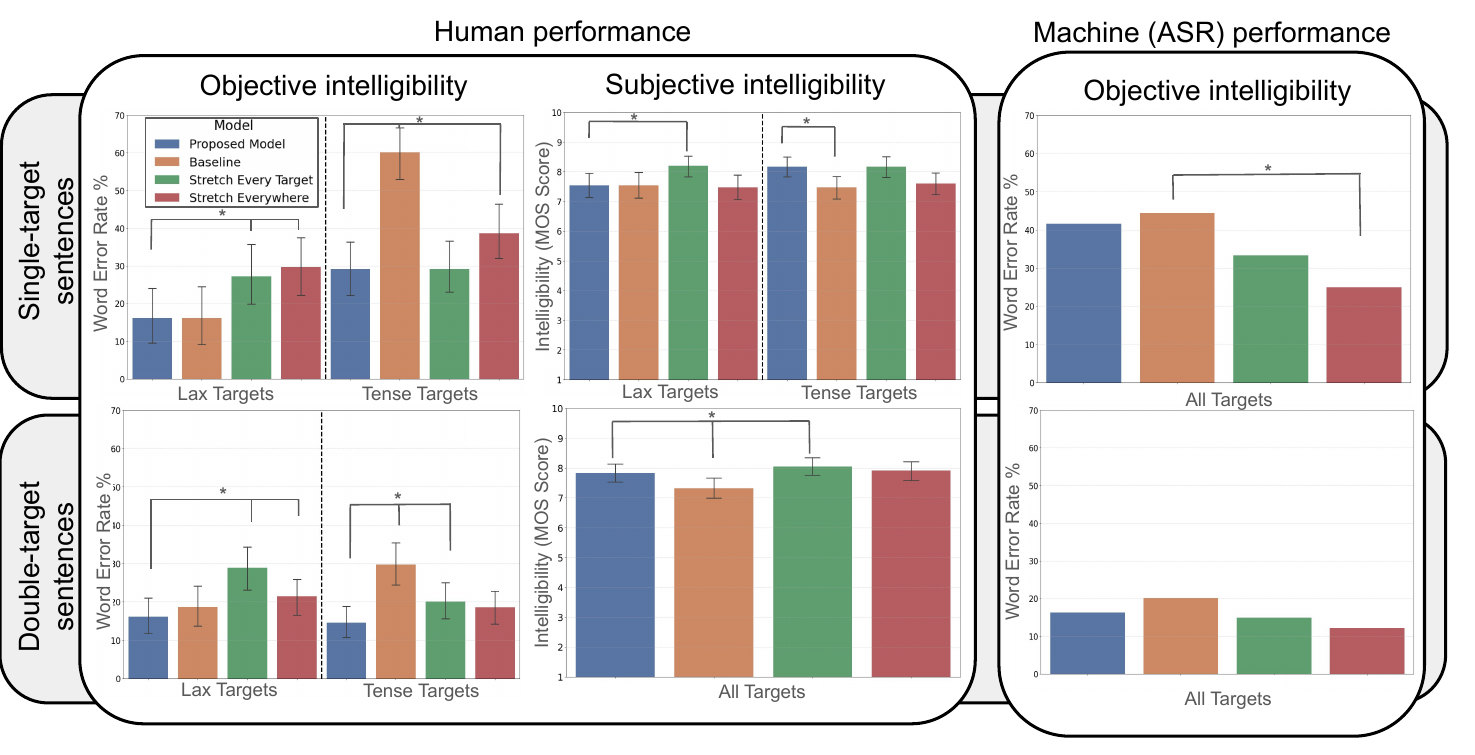}
  \caption{{\bf Speech rate manipulations improve word comprehension of machine speech for non-native listeners, although they remained unaware of the facilitating effect}. {\bf Left:} Objective intelligibility (word error rate) for 4 alternative strategies of speech-rate ajustments, measured on a sample of N=56 French-L1 participants listening to a set of 32 test sentences. Our proposed data-driven strategy (proximal slowing-down for tense vowels, no change for lax vowels) significantly reduced WER compared to baseline and other global slow-down strategies, both for sentences containing a single target word ({\bf Top}) and double target words ({\bf Bottom}). {\bf Middle:} Subjective intelligibility (MOS score) measured in the same task. Listeners rated models that stretched lax targets as significantly more intelligible, in both single-target ({\bf Top}) and double-target sentences ({\bf Bottom}), although these strategies actually increased WER. {\bf Right:} Automatic machine speech recognition (ASR) word error rate on the same task. For machines contrary to humans, the `stretch-everywhere' strategy significantly improved WER over baseline.}
  \label{fig:wer_tts}
  %\vspace*{-5mm}
\end{figure*}

\subsubsection*{Study 3a: Speech rate provides a powerful strategy to improve word comprehension of machine speech for non-native listeners.}

We tested the effect of that generative model on a large set of 32 sentences, containing either single (e.g. \emph{I saw \ul{full} written on the note pad}) or double target words (e.g., \emph{In his talk he kept using \underline{could}, but I am pretty sure he meant \underline{cooed}}), and all tense/lax pairs in the standard English inventory: \textipa{/i/} vs \textipa{/I/} (e.g. peel vs  pill), \textipa{/u/} vs \textipa{/U/} (e.g. fool vs full), \textipa{/A/} vs \textipa{/2/} (e.g. cot vs cut). Each sentence was presented in two versions, by permuting the target word in each pair (Supplementary Table 1, and \emph{Methods}). One group of N=56 French-L1 participants was asked to listen to each synthesized sentence and to select which target words they heard as part of a list of four options: the correct word, the word with the opposite vowel in the pair, and 2 distractors. 

We compared our proposed generative model with 3 alternative generative rules: a ``baseline'' model which simply used the pre-trained MatchaTTS durations at 0.75x duration; a ``stretch-everywhere'' model, which applied a 1.2x (1.6x $\times$ 0.75) stretch (25\% slower speech rate) across the entire phrase (aimed to mimic the tendency to overall slow down when addressing non-native speakers; \cite{SMIL09,AOKI24}); and an ``stretch-every-target'' model, which applied 1.2x stretch across all target words (regardless of tense/lax) and 0.75x elsewhere. In comparison, our proposed model applied 1.2x (1.6x $\times$ 0.75) stretch proximally to target words containing a tense vowel, and 0.75x elsewhere. 

For sentences containing a single target word, the baseline word error rate (WER) was $M=38.4\%$, which was largely driven by errors on tense vowels (lax: $M=16.9\%$; tense: $M=60.2\%$), translating a response bias towards lax vowels that was also seen in Study 2. For tense targets, our proposed model (which was the same as ``stretch-every-target'') significantly reduced WER ($M=29.5\%$) compared to both baseline ($M=60.2\%$, a 51\% relative decrease; $Z(51)=14.5$, $p<.001$) and also reduced error rate compared to `stretch-everywhere', although not significantly ($M=37.6\%$; $Z(51)=94.5$, $p=.05$). For lax target words, our proposed model (which was the same as `baseline', $M=16.9$\%) also significantly reduced WER compared to `stretch-everywhere' ($M=29.7\%$, $Z(51)=60.0$, $p=.03$) and ``stretch-every-target'' ($M=28.9\%$, $Z(51)=54.0$, $p=.03$; Figure \ref{fig:wer_tts}-top,left). 

Similar results were found for sentences containing double target words. Baseline WER ($M=24.3\%$) was larger for tense ($M=30.0\%$) than lax targets ($M=18.7\%$, $Z(55)=234.5$, $p<.001$), and was largely reduced with our proposed model ($M=15.2\%$ overall, a $37\%$ relative decrease from baseline). For tense targets, our proposed model significantly reduced WER ($M=14.4\%$) compared to both baseline ($M=29.8\%$, a 48\% relative decrease; $Z(55)=104.5$, $p<.001$) and `stretch-every-target' ($M=20.1\%$, $Z(55)=166.0$, $p=.008$). For lax targets, our proposed model also significantly reduced WER compared to `stretch-everywhere' ($M=21.5\%$, $Z(55)=225$, $p=.02$) and `stretch-every-target' ($M=28.8\%$, $Z(55)=74$, $p<.001$; Figure \ref{fig:wer_tts}-bottom,left).

% Note in double-word sentences, clarity was better than baseline for lax vowels, while they are supposed to be the same. This likely resulted from the phrases containing both a tense and a lax vowel, which means the L2 participants could use duration differences between the two words to more easily differentiate the word. Not described here, but kept for future reference if needed. 

\subsubsection*{Study 3b: Listeners judge the scissor structure less clear than a general slowdown, although the latter actually increases Word Error Rate (WER). }

In addition to word comprehension, participants also rated manipulated sentences on the basis of their naturalness, intelligibility, appropriate prosody, listening effort, respectfulness and encouragement. Strikingly, our results showed that listeners failed to correctly identify our proposed model as optimal for comprehension, and rated models that stretched lax targets as significantly more intelligible than baseline (which was the same as our proposed model) for a single-word (``stretch-every-target'': intelligibility $MOS=8.18$ vs $7.63$, $Z(51)=116$, $p=.002$; Figure \ref{fig:wer_tts}-top,middle), although these strategies actually increased WER. Similar results were found in double-word sentences (``stretch-every-target'': intelligibility $MOS=8.06$ vs $7.83$, $Z(55)=375.5$, $p=.01$; Figure \ref{fig:wer_tts}-bottom,middle). In short, participants declarative knowledge on what constitutes clear speech contradicted what actually helped them comprehend speech. 

In addition, the common clear-speech strategy of ``stretching-everywhere'' was also considered less natural, less prosodically appropriate, less respectful and less encouraging than baseline and other more targeted strategies (Supplementary Table 2), even in cases where it showed objective improvements of WER. This suggests that maintaining natural speech rate outside of difficult words can make the L2 listeners feel less patronized. 

\subsubsection*{Study 3c: Automatic machine speech recognition does not use the same duration mechanisms as humans when facing difficult words.}

Finally, we submitted the same set of 32 test sentences (plus another 41 constructed with the same procedure -- see \emph{Methods}) to a state-of-art machine-learning speech recognizer (Whisper ASR, \cite{RAD23}), and tested the effect of the four models on the machine's WER. Baseline WER (combined over single-word and double-word sentences) was 21.4\% for the recognition of target words (same task as human participants), and 17.1\% for the recognition of the complete sentences. This performance is worse than the expected 5-10\% WER for the Whisper algorithm, probably because the test material does not provide contextual cues \cite{LALL24}. Strikingly though, we did not see the same improvements in WER with our proposed scissor model that we did for L2 listeners: for machine ASR, `stretch-everywhere' improved WER the most (targets: $16.4\%$, complete sentences: $15.98\%$), while our proposed model barely outperformed baseline (Fig. \ref{fig:wer_tts}-right). The difference between `stretch-everywhere' and baseline was significant in the case of single-target sentences ($\chi^2$(1, N = 36) = 4.11, p = .043). In addition, all three rate manipulations increased the proportion of out-of-pair errors, i.e. hearing `peel` as `peak` instead of `pill`, compared to baseline (baseline: 28.6\% out-of-pair; `stretch-everywhere': $68.4\%$; `stretch-every-target': $77.3\%$; our model: $70.8\%$). In sum, state-of-art machine ASR does not rely on the same duration mechanisms as humans when facing ambiguous pairs of words. 

\section*{Discussion}

While human speakers often use globally-slower speech to address listeners with perception challenges, the ability of machine-generated speech to produce targeted slowing of hard-to-comprehend speech sequences promises enhancements over what humans naturally produce. Here, we reported the first study to systematically identify how targeted speech-rate adjustments can improve L2 listeners’ perception, and did so by constructing a data-driven algorithm to strategically manipulate speech rate in machine-generated speech. In Study 1, we showed that the temporal influence of speech rate prior to a target vowel contrast is remarkably stable across both L1 and L2 English listeners (French, Mandarin and Japanese). In Study 2, we then showed that this scissor-like structure not only had a large facilitating effect on how L2 listeners comprehend English vowel contrasts, but that L1 listeners also relied on this pattern as a fall-back strategy in difficult acoustic conditions. Finally, in Study 3, we modified a state-of-the-art speech-synthesis algorithm to enforce this temporal structure on arbitrary speech and found that, strikingly, listeners remained unaware of the facilitating effect of targeted speech rate on word comprehension: on the contrary, they judged global slowing as more intelligible even though it actually increased comprehension error. Taken together, these results show that targeted changes to speech rate can aid listeners’ speech intelligibility under challenging listening conditions, but that subjective impressions of this mechanism diverge from their actual perceptual efficacy.  

While our study is by no means the first to uncover contextual speech rate effects on the perception of vowels and words \cite{MillerLiberman1979, KIDD89,SommersNygaardPisoni1994, NEWM96,STIL20}, it is, to the best of our knowledge, the first to examine the question using a data-driven reverse correlation paradigm (for a related recent reverse-correlation study on how prosodic cues influence the perception of word boundaries, see \cite{OSS23}). Our Study 1a of French and English isolated words (Suppl. Fig. 1) showed that the reverse correlation procedure can uncover speech rate characteristics of vowels in widely generalizable situations, covering both phonetically-predictable (English) and undocumented cases (French). For the English tense-lax contrast \textit{peel} versus \textit{pill} where duration is a secondary cue \cite{SOLE10}, we recovered a slower rate for \textipa{/pIl/} vs \textipa{/pil/} as predicted phonetically \cite{KOND08}. For the French high-front \textipa{/pyl/} vs high-back and rounded \textipa{/pul/} contrast, in which duration isn't a primary cue, we uncovered that lengthening the sound causally biased perception towards the latter. That effect in French is possibly linked to the presence of the consonant \textipa{/l/} terminating the syllable and may be explained by perceptual expectations of the additional time required for the articulators to move from the back vowel to the relatively front position of the following consonant (see e.g. the 70\% duration difference found between \textipa{/yl/} and \textipa{/al/} in \cite{OSHAU81}). While here we only use these results on isolated words as a baseline for subsequent studies at the sentence level, we believe they already are testimony to how reverse-correlation procedures can benefit phonetic research, providing an interesting data-driven alternative to hypothesis-driven paradigms based on phonetic predictions (see SI Text 1 for an exploratory analysis of pitch cues in Study 1).  

Study 1b of French and English sentences (Figure \ref{fig:revcor}) provided a detailed account of how contextual speech rate may bias the perception of vowels when they're embedded in a phrase. We were able to replicate the contrastive context effects widely documented in the speech perception literature (for a review, see \cite{STIL20}). Reverse correlation, however, helped us to uncover unprecedently fine-grained information on these effects. In particular, we found that the effect is distally contrastive (slower context makes vowels appear shorter), but becomes congruent proximal to the word, with a tipping point around 200-300ms pre-target, creating a temporal contour which we described as ``scissor-shaped''. Within these kernels, our experiment also allowed quantifying the respective perceptual strength of rate-information intake at various time points in the sentence. While we only used a single leading phrasal context (English: \emph{``I heard them say XXX''}; French: \emph{``Je l’ai entendu dire XXX”}, and a fixed temporal resolution of 100ms (see Material and Methods), further work could investigate how kernels, including their distal/proximal cutoffs, depend on the fine morphological structure of the sentence, such as co-articulation.  

The scissor-shaped kernels were strikingly preserved across both L1-English and L2-English listeners with French, Mandarin and Japanese as their L1s (Study 1c), even though these 4 languages widely differ in how they use duration. While English does not have vowels distinguished primarily by duration, it has vowel pairs (such as the tense-lax contrast investigated here) that are distinguished by duration in conjunction with other spectral information \cite{PET60}. In contrast, vowel duration plays little role in distinguishing vowels in Parisian French \cite{STRAN07} %(although it may in other dialects \cite{MILLER11}) 
and in Mandarin - to the point that Mandarin speakers display reduced precision in determining vowel length in other languages \cite{LU21}. Conversely, all 5 vowels in Japanese contain a minimal pair of short long vowels, which, when varied, can change the meaning of a word \cite{HIRATA2004565}. It is therefore striking that, despite this linguistic variability, all three English-L2 populations relied on contextual speech-rate in the same manner as English-L1 listeners.

While it is well-documented that non-native listeners rely strongly on local (i.e. within-vowel) duration cues to discriminate tense-lax English vowels \cite{KOND08}, prior evidence for duration effects in the preceding context in L2 listeners is scarce and inconsistent. In \cite{BOSK17}, German and Dutch-L1 participants were tested in pseudowords embedded in sentences of the other language. The effect of contextual speech rate on the perception of foreign \textipa{/a/} vs \textipa{/a:/} pairs was preserved in German but not Dutch listeners (where it was reversed, maybe because of competing spectral cues). In general, studying speech-rate effects in L2 participants is almost always construed as testing its 'preservation' in difficult listening conditions, such as cognitive load \cite{BOSK17-2} or acoustic noise \cite{REIN22}, with the expectation that these effects are reduced. In contrast, our results, both at the level of kernels (Study 1) and subsequent validation in noise for L1 speakers (Study 2) indicate that the clarifying effect of speech rate on vowel contrasts actually \emph{increases} with difficult conditions, and in a way that is largely immune to allophonic experience. 

Study 2 extends these results by quantifying the impact of speech-rate kernels on word recognition accuracy when, unlike Study 1, words were not manipulated to have ambiguous formants. In English-L2 listeners (French, Mandarin, Japanese), the effect of speech rate manipulations was very large, reaching an absolute 20-30\% increase in accuracy. That effect size is equivalent to the difference between L1 and L2 speakers at the baseline speech rate. In contrast, English-L1 participants remained remarkably immune to the speech-rate manipulations. Strikingly though, English-L1 participants started relying on speech rate in the anecdotal case of one badly-synthesized sound (\emph{fool}, at very low baseline accuracy 32\%) and, as tested experimentally, with the addition of acoustic perturbation. This pattern of results suggests the existence of a generic timing-perception mechanism which is used principally by L2 participants and which native speakers, who can easily use spectral cues, will only fall-back to in challenging listening conditions. This mechanism is compatible with the desensitization hypothesis \cite{BOHN95} which posits that adult second-language learners become less sensitive to non-native phonetic contrasts due to increased reliance on their first language’s phonological system, rather than simply transferring first-language perception patterns. Furthermore, independent of the vowel system in their native language, listeners may also have duration more readily available as a global speech cue \cite{CHLAD13}. 

This `trading relation' \cite{GOTT88} between temporal and spectral information provides a striking example of cue weighting \cite{SCH20}, the ability to integrate and weight information across multiple acoustic dimension in speech perception and production. Cue weighting is known to depend on language \cite{GOTT88,KOND08}, dialect \cite{ESCU04} and L2-learning \cite{MORR02}. For instance, American English-L1 participants classify spectrally-ambiguous French vowels more often as \textipa{/o/} when these vowels were long and as \textipa{/0/} when short, while native French listeners show little effect of duration \cite{GOTT88}. Similarly, Mandarin learners of English use duration more than spectral information for the American English \textipa{/I/} vs \textipa{/i/} contrast, Spanish listeners use both dimensions equally, and American English listeners have a preference for the spectral cues \cite{FLEGE97}. Our results with degraded conditions show that English-L1 participants are able to flexibly adapt cue weights to contextual demands, such as acoustic noise or suboptimal speech synthesis, which make their preferred cue (spectral) more difficult to process or less reliable. This capacity for reweighting is reminiscent, e.g., of Mandarin Chinese listeners swiftly transitioning from using pitch to duration and amplitude cues to recognize lexical tone in whispered speech, which lacks the primary cue of pitch \cite{ZHAN22} and, on a longer developmental scale, on older adults adapting to deficits in auditory temporal discrimination with an increased reliance on spectral cues for voicing judgments (\textipa{/b/} vs. \textipa{/p/}) \cite{TOSC19}.  
Finally, in Study 3, we modified a state-of-art speech-synthesis algorithm to automatically parse text for difficult English vowel contrasts and enforce our previously-learned changes of speech rate in order to facilitate comprehension by English-L2 participants. Implementing the mechanisms identified in Study 1-2 in a full-fledge generative algorithm provides a particularly broad causal test for how targeted changes of speech rate affect intelligibility. Our proposed model significantly reduced WER not only with respect to baseline, but most strikingly, to other typical slowdown strategies used by speakers to address L2 listeners. We note in passing that, even though the scissor-shape pattern of Study 1 was modified in Study 3 to only include proximal context, the fact that this algorithm yielded a lower WER than a global slow-down is consistent with the perceptual importance of a distal contrast of speech rate. 

The fact that listeners judge the targeted speech-rate structure subjectively less intelligible than a general slowdown, although the latter actually increases WER, is interesting with respect to recent research in non-native directed speech (NNDS). When produced by L1 speakers, foreigner-directed speech massively features a reduction of speech rate \cite{AOKI24}, as well as an omnibus lengthening and hyperarticulation of vowels \cite{PIAZ25}. While these strategies are typically evaluated subjectively as more intelligible by L2-participants \cite{BOBB19}, there is debate whether they in fact improve objective word comprehension \cite{PIAZ23,AOKI24}. Consistent with our results, work in \cite{AOKI24} found that NNDS provided no intelligibility advantage over L1-California-English casual speech for Mandarin-L1, English-L2 participants (although hard-of-hearing directed speech, which involves additional spectral adaptations, provided some advantage for L2). More generally, the separation of objective and subjective assessment of intelligibility in our L2 participants is testimony of the difficult conscious/declarative access to linguistic competence in L2 language learning \cite{ELL05}, and is also reminiscent of the mismatch between the conceptual belief of what prosodic cues are used to recognize certainty in speech, and what these cues actually are \cite{Goupil21}. 

Finally, the fact that, contrary to our L2-participants, a state-of-art ASR algorithm did not appear sensitive to the same speech-rate manipulations, preferring a global slowdown or even no slow-down at all, is an interesting case of dissociation of performance between machines and humans. A wealth of recent studies has demonstrated that, while humans and machines may demonstrate ``superficial similarities'' of overall accuracy (e.g., 5-10\% typical WER for Whisper ASR), they often display ``deep differences'' \cite{NERI22}, revealed by different sensitivity to abnormal or adversarial examples \cite{ELSA18,LEPO22} or processing biases (e.g. shape vs texture in face recognition, \cite{DAU21}). It is therefore possible that the performance mismatch between human listeners and machine is the `signature' of an underlying difference of cognitive architecture \cite{FIRE20} in how the Whisper ASR algorithm processes temporal information such as speech rate, perhaps because of how the Transformer architecture was trained to allocate `self-attention' \cite{AB20} across tokens. However, it is also possible that architectures such as Whisper have the cognitive capacity to learn and use human-like patterns of speech-rate integration, but that they weren't trained on datasets including substantial changes of speech-rate and were therefore unable to learn how to combine this information with spectral cues. Further work based on explainable AI approaches could clarify the `performance vs competence' nature of this performance mismatch \cite{LIN24}. In any case, the fact that Whisper ASR does not scale intelligibility/word-error-rate in the same manner as human listeners when speech rate is varied should prompt caution when using ASR as a proxy to automatically evaluate the intelligibility of speech synthesis algorithms, which is a common practice in the speech technology community \cite{KARB22}. 

Taken together, this work reveals the precise temporal structure by which acoustic context causally shapes speech perception. Doing so, it advances an explicit `algorithmic theory' of intelligibility and demonstrates how covert, data-driven adaptations of speech timing can improve the accessibility of machine-generated speech across diverse listeners and environments. We propose that the same method be generalized to optimize speech synthesis for the comprehension of specific groups of L2 speakers, young or elderly users \cite{TOSC19}, or patients with specific hearing impairments or language-cognitive signatures \cite{BIND25}. 

\footnotesize
\section{Materials and Methods}

\subsection*{Study 1} {\bf Participants: } N=160 participants took part in the study: N=60 English-L1 speakers (female: 33, M=30.0yo ± 10.7, primarily from anglophone Canada), N=54 French-L1 speakers (female: 24, M=32.4yo ± 9.7, primarily from France), N=30 Mandarin-L1 speakers (female: 15, M=22.2yo ± 5.6, primarily from China), and N=16 Japanese-L1 speakers (female: 5, M=29.9yo ± 12.1, primarily from Japan). Participants were recruited via the online Prolific platform. English-L2 participants had a self-rated English proficiency ranging from 2-5 (1: no proficiency, 5: fluent) with a mode of 4. All participants provided their informed consent and were compensated financially for their time at a standard rate. The procedure was approved by the SFU-REB. {\bf Word generation: } We selected two word pairs in English and French that include vowel contrasts known from the literature to be difficult for L2 speakers (for a review, see {\bf SI Text 2}): the tense/lax contrast ``peel'' \textipa{/pil/} and ``pill'' \textipa{/pIl/} in English, and the front/back contrast  ``pull'' \textipa{/pyl/} and ``poule'' \textipa{/pul/} in French. We generated stimuli using the CoquiXTTS text-to-speech algorithm (\url{https://huggingface.co/spaces/coqui}). TTS language was set to English for English stimuli, and French for the French stimuli. An L1 male reference voice was provided, and no other modifications were made to the TTS settings. Word stimuli were generated as part of the phrase \emph{`` I heard them say X''} (French: \emph{``Je l'ai entendu dire X''}), then cut to constitute isolated words. {\bf Formant manipulation} We then transformed each synthesized word pair to generate a single morphed sound that was perceptually intermediate between each vowel. To do so, we used the Praat software \cite{praat} to extract the average frequency of each original word's F1 and F2 formants (\textipa{/i/}: F1= 305.89Hz, F2= 2440.77Hz; \textipa{/I/}: F1= 476.85Hz, F2= 1565.45Hz; \textipa{/u/}: F1= 238.90Hz, F2= 1005.67Hz; \textipa{/y/}: F1= 212.61Hz, F2= 1519.49Hz), as well as transform these values towards one another, within the pair, on a 10-Hz step-size grid. We then evaluated the phonetic ambiguity of each transformed sound on the grid by computing the log-probability of recognizing either target word using the Whisper speech recognition software \cite{RAD23}, and selected the F1,F2 transformation with the minimum average difference (in log probability) to p=0.5. This procedure provided two word stimuli, one averaged between \textipa{/i/}-\textipa{/I/} (F1= 436.77Hz, F2= 1722.68Hz, transformed from \emph{peel}), and the other between \textipa{/u/}-\textipa{/y/} (F1= 238.13Hz, F2= 1258.68Hz, transformed from \emph{poule}). {\bf Phrase generation} Transformed stimuli were either presented as isolated words (word conditions), or as part of a phrase (phrase condition). For the phrase conditions, transformed stimuli were inserted back into the original sentence, at a manually-selected zero-crossing 120ms after the end of the last word \emph{say/dire}. {\bf Speech rate manipulations} Finally, we used the CLEESE software \cite{BURR19} to systematically randomize the pitch and speech-rate of the word and phrase stimuli, preparing them for reverse-correlation. While pitch was not our primary object of study, pitch transformations were introduced to increase the prosodic variability of stimuli and decrease experimental demand with respect to speech rate manipulations. The pitch contour of the stimuli was first algorithmically flattened to a constant 120Hz. The stimuli were then cut into successive 100ms windows (\emph{n}=4 for words; \emph{n}=13 for phrases). Each window was transformed by pitch shifting with a factor normally distributed with $\mu$=0, $\sigma$=100 cents (i.e., $\pm$ 1 semitone), then by time-stretching with a factor normally distributed with $\mu$=0\%$,\sigma$=100\% (i.e. doubling or halving the window's duration; both distributions clipped at $\pm 2 \sigma$ to avoid extreme transformation values). We generated 500 random transformations of each stimulus. \noindent{\bf Procedure: }The pool of participants took part in four experimental conditions, which differed by language and stimulus type (English or French stimuli, word or phrase conditions). In each condition, participants were presented with a series of 250 1-interval, 2-alternative forced choice trials (1-I, 2-AFC), each consisting of one random pitch and stretch transformation of the stimulus. For each trial, participants were asked which of the two target words they heard. Each condition lasted on average 15min. All four conditions were carried out by N=25 English-L1 and French-L1 participants. In addition, the two English-stimulus word and phrase conditions were also carried out (each) by N=25 Mandarin-L1 participants, as well as a smaller sample of Japanese-L1 participants (words: N=13; phrases: N=14) due to recruitment difficulties. The majority of participants took part in one single (randomized) condition. A small proportion participated in more than one, in which case the order was randomized across language and stimulus type. Data from each condition was analyzed as independent regardless of repeated measures. \noindent {\bf Reverse correlation analysis: }Reverse correlation allows us to infer which stimulus features systematically bias listeners’ perceptual decisions, by analysing the stimulus variations that precede each response. In the present study, this approach was used to estimate how speech-rate changes at different moments in time influenced listeners’ lexical decisions. We used the {\it classification image} method \cite{Mur11} to compute reverse-correlation kernels for each participant. Each kernel can be interpreted as a temporal weighting function, describing, at each 100 ms time step, whether locally slowing or speeding the speech signal biased the participant’s response toward one lexical alternative or the other. To compute these kernels, we computed the average transformation vector for all trials categorized as each alternative, and then normalized the resulting kernel by dividing it by the root-mean-square sum of its values. This procedure isolates the consistent temporal patterns of rate variation that were causally associated with listeners’ perceptual choices. While our primary focus was speech rate, a similar analysis was conducted, in an exploratory manner, using pitch transformations - see {\bf SI Text1}. {\bf Statistical analysis : } In each condition and sample of participants, we compared the kernels of the two alternative words using paired t-tests over participants. The tests were performed independently at every 100ms time-step, with no correction for multiple measures. 

\subsection*{Study 2} {\bf Participants: } N=145 participants took part in the study: N=50 French-L1 speakers (female: 24, M=34.5yo$\pm$11, primarily from France), N=50 English-L1 speakers (female: 17, M=34.3yo$\pm$9.7, primarily from Canada), N=10 Mandarin-L1 speakers (female: 7, M=28.7yo$\pm$7.8, primarily from China) and N=10 Japanese-L1 speakers (female: 8, M=36.4yo$\pm$13.4, primarily from Japan). Participants were recruited via the online Prolific platform. French-L1 participants had a self-rated English proficiency ranging from 1-5 (1: no proficiency, 5: fluent) with a mode of 5, Mandarin-L1 participants ranging from 2-5 with a mode of 4, and Japanese-L1 participants ranging from 3-5 with a mode of 5. All participants provided their informed consent and were compensated financially for their time at a standard rate. The procedure was approved by the SFU-REB. {\bf Word selection:} The English vowel pair ``pill'' and ``peel'' was again used with the addition of a second tense-lax pair: `full'' (\textipa{/fUl/}) and ``fool'' (\textipa{/ful/}), inserted as above in the sentence ``I heard them say XXX''. {\bf Stimulus generation: } The stimuli were generated using Matcha-TTS \cite{mehta2024matcha} (single-speaker model, LJ Speech checkpoint), which has phoneme level duration control via a duration array of the same length as the phonemized phrase. All stimuli were generated as 0.75x the baseline Matcha-TTS speed. In addition, we controlled the duration of the generated stimuli so that it followed the ``scissor-shape'' profile of speech-rate uncovered in Study 1, at 11 levels of positive and negative (i.e., reversed) intensity. To do so, we simplified the kernel to a piecewise linear shape, including only two stretch factors: one constant stretch spanning all the context sentence until the pause before the target word, and another constant stretch spanning the target word. The context stretch was varied up from 0.67x (slower) to 1.5x (faster) speed at increments of 0.1, and the word stretch was varied simultaneously and respectively from 2.0x (faster) speed to 0.5x (slower) at increments of 0.2 (see Fig. \ref{fig:validation_1}). This procedure resulted in 44 different stimuli (4 phrases $\times$ 11 manipulations). {\bf Noise generation} For the ``background noise'' condition (Fig. \ref{fig:validation_noise}), we manipulated the above stimuli using the Audacity software (https://www.audacityteam.org): first, we applied full-wave rectifier distortion (wet/dry: 45\%); then reverberation (room size: 22\%; pre-delay: 10ms; reverberance, dampening: 50\%, wet/dry: -1bB). We then mixed the stimuli with background crowd noise (\url{https://pixabay.com/sound-effects/search/crowd/}) at a manually-determined level to mask spectral content without hindering duration perception (the noise level was the same for every stretch factor). The resulting stimuli was reminiscent of a loud-speaker announcement in a public space: distorted, distanced, with crowd sound in the background.  {\bf Procedure: } The task was the same as Study 1 (1-interval, 2-alternative forced choice). Each of the 44 stimuli (4 phrases, 11 manipulations) was presented 5 times, in random order, resulting in 220 trials (c. 30 min). All 4 participant groups (N=25 for English-L1 and French-L1, N=10 for Mandarin-L1 and Japanese-L1) took part in noiseless conditions. One additional English-L1 group (N=25) took part in the noise condition (Fig. \ref{fig:validation_noise}, and one French-L1 group (N=25) took part in a variant of the experiment were the stretch was only applied to the word, as a control for Study 3 (Suppl. Fig 2). 
{\bf Statistical analysis : } For each target word and participant, we normalized the accuracy (proportion of correct responses) at each stretch factor, relative to the participant's baseline accuracy at 1.0x. We then tested significance at each stretch factor, over the group of participants, using one-sample t-tests against baseline (0), corrected for multiple comparisons (n=10) with the Holm-Bonferroni method. 

\subsection*{Study 3} {\bf Participants:} N=56 French-L1 participants, primarily from France, took part in the study (female: 29, M=34.6yo$\pm$10.7). Participants were recruited via the online Prolific platform. The participants' reported English 1-5 proficiency ranged 2-5, with a mode of 5 (N=29). All participants provided their informed consent and were compensated financially for their time at a standard rate. The procedure was approved by the SFU-REB. {\bf Modified Matcha-TTS algorithm: } We modified the Matcha-TTS algorithm \cite{mehta2024matcha} to parse the text to be synthesized for user- or machine-flagged difficult words (using exclamation points, e.g. ``!peel!''). The baseline speech rate for Matcha-TTS was set at 0.75x duration (i.e. slighly faster than default). For each word, we used the open-source software eSpeak NG to extract a phonemized array, and checked if it contained tense or lax vowels, as well as primary stress. If the word contains tense vowels and no lax vowels, we applied a 1.2x (1.6x $times$ 0.75x) stretch factor across the word, as well as a linear ramp-up and down to 0.75x over the 6 phonemized items (if there are 6 phonemized items between target words, otherwise as many as are available) that preceded and followed the word (amounting to 200-300ms identified in Study 1). If the word contained only lax vowels, no modification was applied (i.e. stretch remained at 0.75x). Words contained both tense and lax vowels were treated as tense if the tense vowel had primary stress (e.g. music \textipa{/\textprimstress mjuzIk/}). {\bf Control algorithms: } We compared our proposed model (above) with 3 alternative algorithms: \emph{Baseline}: the base Matcha-TTS, synthesized at 0.75x speech rate; \emph{Stretch-everywhere}: 1.2x ($0.75\times1.6)$ speech rate applied across the entire phrase; \emph{Stretch-every-target}: 1.2x stretch applied across all target words, 0.75x speech rate elsewhere, regardless of content. All were all generated using Matcha-TTS as above. {\bf Word selection}: We tested the following English vowel contrasts and word pairs : \textipa{/i/} vs \textipa{/I/} (peel vs pill, scene vs sin, sheep vs ship, beat vs bit, bean vs bin, keyed vs kid, reap vs rip); \textipa{/u/} vs \textipa{/U/} (fool vs full, cooed vs could, pool vs pull, wooed vs wood, shooed vs should); \textipa{/A/} vs \textipa{/2/} (cot vs cut, knot vs nut, hot vs hut, bought vs but, cop vs cup, doll vs dull). Words were inserted either as a single target in a semantically-neutral phrase (8 phrases, e.g. \emph{``she kept mentioning \ul{cot} during the conversation''}) or in phrases containing two targets (8 phrases, e.g. \emph{``in his talk he kept using \ul{could}, but I'm pretty sure he meant \ul{cooed}''}), which were either (1) a tense and a lax minimal pair, (2) a tense and a lax that are not minimal pair, (3) two tense, and (4) two lax vowels (Supplementary Table 1). This resulted in 16 stimulus phrases (which were presented manipulated by each of the 4 alternative algorithms, see below). In addition, we created 16 additional distractor phrases (to be presented only once, see below), which were the same as each stimulus phrase by replaced each target word with the opposite word in the pair (e.g. \emph{``she kept mentioning [\ul{cot} $\rightarrow$ \ul{cut}] during the conversation''}). {\bf Procedure:} Participants were presented with trials consisting of one synthesized sentence, as well as a written transcription with target words removed (e.g. \emph{``in his talk he kept using X, but I'm pretty sure he meant Y''}). They had to select which target words they heard from a list of 4 options which included the correct word (e.g., could), the opposite word in the pair (e.g., cooed) and 2 distractors: same word with another vowel (e.g., cod), and a dissimilar word (e.g., shop). In two-word sentences, participants were instructed that it was possible to hear the same word twice (although it was never actually the case). Each stimulus phrase was presented four times, as generated by our proposed algorithm (proximal slow-down on tense vowels) and the three alternative algorithms (baseline, stretch-everywhere, stretch-every-target). In addition, distractor phrases were presented only once, in one random generation algorithm. This resulted in 80 trials (16 stimulus phrase x 4 + 16 distractor phrases), which were presented in a random order within single-word and double-word phrase blocks, also randomized. For each trial, after participants selected the word(s), they also responded to a mean-opinion-score (MOS) questionnaire (0-10 Likert scales) assessing the sentence's perceived naturalness (\emph{``How natural (pleasantly humanlike) was the sound of the voice''}, anchored by ``Extremely unnatural'' and ``Perfectly natural''), intelligibility (\emph{``Please rate the extent to which it was easy or difficult to understand what the voice was saying''}, anchored by ``Completely unintelligible'' and ``Completely intelligible''), prosody (\emph{``To what extent were the elements of timing, pitch and emphasis appropriate for the message''}, anchored by ``Completely inappropriate'' and ``Always appropriate'') taken from the MOS-X2 set \cite{LEW18}; listening effort (\emph{``Please rate the degree of effort you had to make to understand the message''}, anchored by ``Impossible even with much effort'' and ``Not effort required'') taken from the MOS-X set \cite{LEW18}; respectfulness (\emph{``For an English second language speaker being spoken to with this voice, how respectful is the voice?''}, anchored by ``Condescending'' and ``Respectful'') and encouragement (\emph{``For an English second language speaker being spoken to with this voice, how encouraging is the voice?''}, anchored by ``Not encouraging'' and ``Encouraging'') taken from L2-directed speech research \cite{ROTHERMICH201922}. The experiment was administered on the web, and lasted approximately 20 minutes. {\bf Machine perception:} Whisper ASR \cite{RAD23} was used to conduct the same experiment. The stimuli consisted of the same 32 sentences used for human experiments, and an additional set of 41 sentences constructed on the same model. Stimuli were fed to the ASR system which provided a transcription of the sentence, and WER on the sentence's target words was computed in \% of presented sentences. {\bf Statistical analysis : } For human participants, the analysis of WER was separated into single, and double target word phrase sets, as well as tense and lax vowel categories were appropriate. In each TTS condition (ours, baseline, stretch-everywhere, stretch-every-target),  WERs were calculated as the percentage of correctly selected responses across the 16 stimulus sentences. Because of non-normality, WER and MOS scores were compared across conditions using Wilcoxon-signed rank tests, with Holm corrections for multiple comparisons over conditions. For Whisper ASR, WER was calculated on target words separately over single-target sentence and double-target sentence. The proportion of correct and incorrect answers was compared across conditions using a $\chi^2$ test.

\section{Acknowledgements}

Work funded by Agence Nationale de la Recherche Sounds4Coma (ANR-21-CE19-0048) and Fondation pour l'Audition DASHES (FPA RD-2021-12)  (to JJA), conducted in the framework of the EIPHI Graduate school (ANR-17-EURE-0002 contract), the Simon Fraser University FASS Breaking Barriers Interdisciplinary Incentive Grant, the France Canada Research Fund, the Mitacs Globalink Research Award, the Canada CIFAR AI Chairs program, the NSERC Discovery Grant (RGPIN-2024-06519). The authors thank P. Maublanc, R. Guha, and A. Adl Zarrabi for their valuable discussions; V. Yang, B. Burkanova, C. Zhang, B.Kwan, F.Wang and M. Durana for their help running our study; and the Rajan Family for their support. 

%\showacknow{} % Display the acknowledgments section

%\bibsplit[12]
%Use \bibsplit to split the references from the body of the text. Value "[2]" represents the number of reference in the left column (Note: Please avoid single column figures & tables on this page.)

\renewcommand\refname{References}
%%%%%%%%%%%%%%%%%%%%%%%%%%%%%%%%%%%%%%%%%%%%%%%%%%%%%%%%%%%%%%%%%%%%%
\begin{footnotesize}

\end{footnotesize}

%\end{linenumbers}
\end{doublespacing}
\end{document}